# A Transformational Characterization of Equivalent Bayesian Network Structures

**David Maxwell Chickering**
Computer Science Department
University of California at Los Angeles
*dmax@cs.ucla.edu*

## Abstract

We present a simple characterization of equivalent Bayesian network structures based on local transformations. The significance of the characterization is twofold. First, we are able to easily prove several new invariant properties of theoretical interest for equivalent structures. Second, we use the characterization to derive an efficient algorithm that identifies all of the compelled edges in a structure. Compelled edge identification is of particular importance for learning Bayesian network structures from data because these edges indicate causal relationships when certain assumptions hold.

## 1 INTRODUCTION

A Bayesian network for a set of variables $U = \{x_1, \ldots, x_n\}$ represents a joint probability distribution over those variables. It consists of (1) a network structure that encodes assertions of conditional independence in the distribution and (2) a set of conditional probability distributions corresponding to that structure. The network structure is an acyclic directed graph (*dag* for short) such that each variable $x_i$ in $U$ has a corresponding node $x_i$ in the structure.

For any given network structure, there is a corresponding set of probability distributions that can be represented using a Bayesian network with that structure. Two network structures are *equivalent* if the set of distributions that can be represented using one of the dags is identical to the set of distributions that can be represented using the other. Because equivalence is reflexive, symmetric, and transitive, the relation defines a set of equivalence classes over network structures.

The notion of equivalence is of particular importance for learning Bayesian networks from data. As we will see in Section 4, *compelled* edges — edges with invariant orientation for all structures in an equivalence class — can indicate causal relationships when certain assumptions hold.

There are two major contributions of this paper. First, we derive a characterization of equivalent network structures based on local transformations. This characterization is presented here fully for the first time, although the result is used in earlier works by Heckerman et al. (1994) and Chickering et al. (1995). The characterization leads to a simple method for proving invariant properties over equivalent structures. We use the method to easily prove that Bayesian networks with equivalent structures require the same number of parameters, and that several well-known scoring metrics used for learning Bayesian network structures from data give the same score to equivalent structures. In addition, we easily prove a graph-theoretic property of equivalent structures that is used by Chickering et al. (1995) to derive an important complexity result.

The second contribution of this paper is the presentation of an efficient algorithm to identify all of the compelled edges in a given network structure. We use the transformational characterization to prove the correctness of the algorithm, and present an implementation of this algorithm that is asymptotically optimal on average.

In Section 2 we describe our notation and introduce previous relevant work. In Section 3 we derive the characterization and prove some invariant properties for equivalent structures. In Section 4 we present a compelled edge identification algorithm and analyze the complexity for various implementations.

## 2 NOTATION

In this section we introduce our notation and discuss previous relevant work on which our characterization is based.

A Bayesian network $B$ is a pair $(\mathcal{G}, \theta_{\mathcal{G}})$ where $\mathcal{G} = (U, E_{\mathcal{G}})$ is a dag, and $\theta_{\mathcal{G}}$ is the set of conditional probability distributions that correspond to $\mathcal{G}$. Throughout this paper, we make many comparisons between dags. It is to be assumed that whenever we make such a comparison, the dags in question are all defined over the same set of vertices $U$, and that the only difference is the set of edges connecting these vertices.

Although we already defined equivalence in Section 1, we present a more formal definition here.

**Definition 1** *Two dags $\mathcal{G}$ and $\mathcal{G}'$ are equivalent if for every Bayesian network $B = (\mathcal{G}, \theta_{\mathcal{G}})$, there exists a Bayesian network $B' = (\mathcal{G}', \theta_{\mathcal{G}'})$ such that $B$ and $B'$ define the same probability distribution, and vice versa.*

We use $\mathcal{G} \approx \mathcal{G}'$ to denote that $\mathcal{G}$ and $\mathcal{G}'$ are equivalent. As was stated earlier, the relation $\approx$ defines a set of equivalence classes over the network structures. We often say that two Bayesian networks are equivalent when we mean that the structures of those networks are equivalent.

A directed edge $x_i \rightarrow x_j \in E_{\mathcal{G}}$ is *compelled* in $\mathcal{G}$ if for every dag $\mathcal{G}' \approx \mathcal{G}$, $x_i \rightarrow x_j \in E_{\mathcal{G}'}$. For any edge $e \in E_{\mathcal{G}}$, if $e$ is not compelled in $\mathcal{G}$, then $e$ is *reversible* in $\mathcal{G}$, that is, there exists some dag $\mathcal{G}'$ equivalent to $\mathcal{G}$ in which $e$ has opposite orientation. We use $C_{\mathcal{G}}$ and $R_{\mathcal{G}}$ to represent the set of compelled and reversible edges in $\mathcal{G}$ respectively. For any node $x_i$ in some dag $\mathcal{G}$, we use $\Pi_i^{\mathcal{G}}$ to denote the set of parents of $x_i$ in $\mathcal{G}$. When the dag $\mathcal{G}$ is clear from context, we use $\Pi_i$ instead. For notational simplicity, we often use symbols without indices for nodes in a dag. In this case, we subscript $\Pi$ with the full name of the node. For example, we use $\Pi_y$ to denote the parent set of node $y$.

The notion of a *covered* edge is very important in the sections to follow, so we emphasize the definition here.

**Definition 2** *An edge $e = x \rightarrow y \in E_{\mathcal{G}}$ is covered in $\mathcal{G}$ if $\Pi_y^{\mathcal{G}} = \Pi_x^{\mathcal{G}} \cup x$.*

In other words, $x \rightarrow y$ is covered in $\mathcal{G}$ if $x$ and $y$ have identical parents in $\mathcal{G}$, with the exception that $x$ is not a parent of itself.

The *skeleton* of any dag is the undirected graph resulting from ignoring the directionality of every edge. For any pair of dags $\mathcal{G}$ and $\mathcal{G}'$ that share the same skeleton, we use $\Delta(\mathcal{G}, \mathcal{G}')$ to denote the set of edges in $\mathcal{G}$ that have opposite orientation in $\mathcal{G}'$. A *v-structure* in dag $\mathcal{G}$ is an ordered triple of nodes $(x, y, z)$ such that (1) $\mathcal{G}$ contains the arcs $x \rightarrow y$ and $z \rightarrow y$, and (2) $x$ and $z$ are not adjacent in $\mathcal{G}$.

The characterization of equivalent structures that we present in the next section is based on that derived by Verma and Pearl (1990) which reads:

**Theorem 1** *[Verma and Pearl, 1990] Two dags are equivalent if and only if they have the same skeletons and the same v-structures.*

A consequence of Theorem 1 is that for any edge $e$ participating in a v-structure in some dag $\mathcal{G}$, if that edge is reversed in some other dag $\mathcal{G}'$, then $\mathcal{G}$ and $\mathcal{G}'$ are not equivalent.

A *topological sort* of the nodes in a dag $\mathcal{G}$ is any total ordering of the nodes such that for any pair of nodes $x_i$ and $x_j$ in $\mathcal{G}$, if $x_i$ is an ancestor of $x_j$, then $x_i$ must precede $x_j$ in the ordering.

# 3   THE CHARACTERIZATION AND SOME CONSEQUENCES

In this section we derive a simple local characterization of equivalent network structures. Simply stated, we show that a property holds for all pairs of equivalent networks that differ by a single covered edge orientation if and only if that property holds for all networks in the equivalence class. In Sections 3.1 to 3.3, we use this new characterization to prove several invariant properties of equivalent structures.

While preparing the final version of this work, we became aware of two previous attempts to prove the correctness of the characterization. Madigan (1993) uses the characterization to prove a property of equivalent structures similar to that of Theorem 1. Bouckaert (1993a) uses the characterization to prove a property of the MDL scoring metric that we explore in Section 3.2. The proof we provide in this section, unlike previous ones, includes the necessary step of demonstrating the existence of an edge that can be reversed and, moreover, explicates constructively how such an edge is found.

As we see in Lemma 1, for any pair of dags that differ by a single edge reversal, there are simple, local, necessary and sufficient conditions for determining if the dags are equivalent.

**Lemma 1** *Let $\mathcal{G}$ be any dag containing the edge $x \rightarrow y$, and let $\mathcal{G}'$ be the directed graph identical to $\mathcal{G}$ except that the edge between $x$ and $y$ in $\mathcal{G}'$ is oriented as $y \rightarrow x$. Then $\mathcal{G}'$ is a dag that is equivalent to $\mathcal{G}$ if and only if $x \rightarrow y$ is a covered edge in $\mathcal{G}$.*

**Proof:**

**(if)** Assume that $x \rightarrow y$ is a covered edge in $\mathcal{G}$. That is, $\Pi_y^{\mathcal{G}} = \Pi_x^{\mathcal{G}} \cup x$. First we show that $\mathcal{G}'$ is a dag. If $\mathcal{G}'$ contained a cycle, then this cycle must include the edge $y \rightarrow x$ because $\mathcal{G}$ is a dag. Thus there must be a directed path from $x$ to $y$ in $\mathcal{G}$ which does not include the edge $x \rightarrow y$. Let $z$ be the last node in this path. By assumption, $z$ is a parent of both $x$ and $y$ in $\mathcal{G}$, and because there is a directed path from $x$ to $z$, this implies that $\mathcal{G}$ contains a cycle, which contradicts the fact that $\mathcal{G}$ is a dag.

Now we show that $\mathcal{G}' \approx \mathcal{G}$. Because $\mathcal{G}$ and $\mathcal{G}'$ have the same skeletons, if $\mathcal{G}' \not\approx \mathcal{G}$ then one of the dags must contain a v-structure that is not in the other. Suppose that $\mathcal{G}'$ contains a v-structure not in $\mathcal{G}$. This v-structure must include the edge $y \rightarrow x$ because this is the only edge that differs between $\mathcal{G}'$ and $\mathcal{G}$. But this implies that $x$ has a parent that is not adjacent to $y$ in both graphs, contradicting the assumption that every parent of $x$ is also a parent of $y$ in $\mathcal{G}$. If we assume that $\mathcal{G}$ contains a v-structure not in $\mathcal{G}'$, a similar argument also yields a contradiction.

(only if) We now show that if $x \rightarrow y$ is not a covered edge in $\mathcal{G}$, then either $\mathcal{G}'$ contains a directed cycle, or $\mathcal{G}'$ is a dag that is not equivalent to $\mathcal{G}$. If $x \rightarrow y$ is not a covered edge in $\mathcal{G}$, at least one of the following two conditions must hold in $\mathcal{G}$: (1) Some node $z \neq x$ is a parent of $y$ but not a parent of $x$. (2) Some node $w$ is a parent of $x$ but not a parent of $y$.

Let $z \neq x$ be a parent of $y$ in $\mathcal{G}$ which is not a parent of $x$ in $\mathcal{G}$. If $z$ and $x$ are not adjacent, then $(x, y, z)$ is a v-structure in $\mathcal{G}$ that does not exist in $\mathcal{G}'$. If $x$ is a parent of $z$ in $\mathcal{G}$, then by definition of $\mathcal{G}'$, it follows that $x$ is a parent of $z$ in $\mathcal{G}'$ and therefore $\mathcal{G}'$ contains a directed cycle.

Let $w$ be a parent of $x$ in $\mathcal{G}$ which is not a parent of $y$ in $\mathcal{G}$. If $w$ and $y$ are not adjacent, then $(w, x, y)$ is a v-structure in $\mathcal{G}'$ that does not exist in $\mathcal{G}$. The node $y$ cannot be a parent of $w$ in $\mathcal{G}$, lest $\mathcal{G}$ would contain a directed cycle. $\square$

Smith (1989) proves the (if) part of Lemma 1 using an additional precondition.

Clearly any property that holds over all dags in an equivalence class must hold over every pair of dags in that class which differ by the orientation of a single covered edge. We prove the converse by showing that for any pair of equivalent dags $\mathcal{G}$ and $\mathcal{G}'$, we can transform $\mathcal{G}$ into $\mathcal{G}'$ by a series of covered edge reversals, where each reversed edge is from $\Delta(\mathcal{G}, \mathcal{G}')$. As we see below, identifying a covered edge in $\Delta(\mathcal{G}, \mathcal{G}')$ is simple.

**Algorithm Find-Edge**$(\mathcal{G}, \mathcal{G}')$
**Input:** Equivalent dags $\mathcal{G}$ and $\mathcal{G}'$ that differ by at least one edge
**Output:** Edge from $\Delta(\mathcal{G}, \mathcal{G}')$
(Let $P_v = \{u | u \rightarrow v \in \Delta(\mathcal{G}, \mathcal{G}')\}$.)

1. Perform a topological sort on the nodes in $\mathcal{G}$
2. Let $y$ be the minimal node with respect to the sort for which $P_y \neq \emptyset$
3. Let $x$ be the maximal node with respect to the sort for which $x \in P_y$
4. Output $x \rightarrow y$

**Lemma 2** *The edge $x \rightarrow y$ output from* **Algorithm Find-Edge**$(\mathcal{G}, \mathcal{G}')$ *is a covered edge.*

**Proof:** Suppose that $x \rightarrow y$ is not a covered edge. Let $z \neq x$ be any parent of $y$ that is not a parent of $x$. If $z$ is not adjacent to $x$ then $x \rightarrow y$ participates in a v-structure in $\mathcal{G}$ that cannot be in $\mathcal{G}'$, contradicting the fact that $\mathcal{G} \approx \mathcal{G}'$. If $x \rightarrow z$ is in $\mathcal{G}$, then either this edge or $z \rightarrow y$ must be in $\Delta(\mathcal{G}, \mathcal{G}')$, lest $\mathcal{G}'$ would contain a directed cycle. If $x \rightarrow z$ is in $\Delta(\mathcal{G}, \mathcal{G}')$ then $z$ would have been chosen instead of $y$ in Step 2. If $z \rightarrow y$ is in $\Delta(\mathcal{G}, \mathcal{G}')$, then $z$ would have been chosen instead of $x$ in Step 3. If we assume that there exists a parent of $x$

that is not a parent of $y$, a similar argument yields a contradiction. $\square$

Using Lemmas 1 and 2 we can prove the characterization.

**Theorem 2** *Let $\mathcal{G}$ and $\mathcal{G}'$ be any pair of dags such that $\mathcal{G} \approx \mathcal{G}'$. There exists a sequence of $|\Delta(\mathcal{G}, \mathcal{G}')|$ distinct edge reversals in $\mathcal{G}$ with the following properties:*

1. *Each edge reversed in $\mathcal{G}$ is a covered edge*

2. *After each reversal, $\mathcal{G}$ is a dag and $\mathcal{G} \approx \mathcal{G}'$*

3. *After all reversals, $\mathcal{G} = \mathcal{G}'$*

**Proof:** We show that all the conditions hold if we use **Procedure Find-Edge** with input $\mathcal{G}$ and $\mathcal{G}'$ to identify the next edge $x \rightarrow y$ to reverse. By Lemma 2, $x \rightarrow y$ is a covered edge and Condition 1 holds. By Lemma 1 and Condition 1, Condition 2 holds. After each reversal $|\Delta(\mathcal{G}, \mathcal{G}')|$ decreases by one and thus Condition 3 holds. $\square$

Using Theorem 2, we can prove that a given property is invariant over all equivalent structures simply by showing that the property is invariant to any reversal of a single covered edge. In the sections to follow, we prove that several theoretically interesting properties are invariant over equivalent structures using this technique.

## 3.1  NUMBER OF PARAMETERS

In this section, we use Theorem 2 to prove that Bayesian networks with equivalent structures require the same number of parameters.

Consider a Bayesian network $B = (\mathcal{G}, \theta_{\mathcal{G}})$ where $\mathcal{G} = (U, E_{\mathcal{G}})$. For any node $x_i \subset U$, we use $r_i$ to be the number of states of $x_i$. For each node $x_i$ with parents $\Pi_i$, $\theta_{\mathcal{G}}$ contains the conditional probability distribution $p(x_i | \Pi_i)$. We use $Dim(x_i, \Pi_i)$ to represent the number of logically independent parameters needed to represent $p(x_i | \Pi_i)$. For every distinct parent instantiation, there are $r_i - 1$ independent parameters, and therefore

$$Dim(x_i, \Pi_i) = (r_i - 1) \prod_{x_j \in \Pi_i} r_j \qquad (1)$$

We use $Dim(\mathcal{G})$ to represent the number of parameters needed to completely specify $\theta_{\mathcal{G}}$. We can express $Dim(\mathcal{G})$ as follows:

$$Dim(\mathcal{G}) = \sum_{x_i} Dim(x_i, \Pi_i)$$

**Theorem 3** *If $\mathcal{G} \approx \mathcal{G}'$ then $Dim(\mathcal{G}) = Dim(\mathcal{G}')$.*

**Proof:** From Theorem 2, we need only show that the theorem holds when $\mathcal{G}$ and $\mathcal{G}'$ differ by the orientation of a single covered edge. Let $x_i \rightarrow x_j$ be this edge in $\mathcal{G}$.

For any node $x_k$, let $\Pi_k$ and $\Pi'_k$ be the parents of node $x_k$ in $\mathcal{G}$ and $\mathcal{G}'$ respectively. Because every node except for $x_i$ and $x_j$ have identical parents in $\mathcal{G}$ and $\mathcal{G}'$, we need only show that

$$Dim(x_i, \Pi_i) + Dim(x_j, \Pi_j) = Dim(x_i, \Pi'_i) + Dim(x_j, \Pi'_j)$$

Plugging Equation 1 into the above expression we have

$$\left[(r_i - 1) \prod_{x_k \in \Pi_i} r_k\right] + \left[(r_j - 1) \prod_{x_k \in \Pi_j} r_k\right] =$$
$$\left[(r_i - 1) \prod_{x_k \in \Pi'_i} r_k\right] + \left[(r_j - 1) \prod_{x_k \in \Pi'_j} r_k\right] \quad (2)$$

By definition of a covered edge, $\Pi_j = \Pi_i \cup x_i$. Furthermore, because $x_i \to x_j$ is the only edge that differs between $\mathcal{G}$ and $\mathcal{G}'$, we have $\Pi'_i = \Pi_i \cup x_j$ and $\Pi'_j = \Pi_i$. After plugging these equalities into Equation 2 and dividing both sides of the resulting equation by $\prod_{x_k \in \Pi_i} r_k$, it is easy to see that the equality holds. $\square$

It follows from Theorem 1 and Theorem 3 that the space needed to represent an arbitrary distribution is identical for all equivalent Bayesian networks. Other consequences of Theorem 3 will be explored in the following section.

## 3.2   SCORE EQUIVALENCE

In this section, we use Theorem 2 to prove that several scoring metrics for learning Bayesian networks from data give the same score to equivalent structures.

A scoring metric is a function that takes as input a Bayesian network structure, a database of observed cases, and possibly some prior knowledge, and returns a value reflecting how well the structure fits the data. We use $\vec{C} = \{C_1, \ldots, C_N\}$ to represent the database of $N$ observed cases and $\xi$ to represent our prior knowledge. We assume that for each case $C_i$, every variable in $U$ is observed.[1]

A metric $M$ is *score equivalent* if and only if

$$\mathcal{G} \approx \mathcal{G}' \Rightarrow M(\mathcal{G}, \vec{C}, \xi) = M(\mathcal{G}', \vec{C}, \xi)$$

for all choices of $\vec{C}$ and $\xi$. When a metric does not use prior information, we omit the argument $\xi$.

For a structure $\mathcal{G}$, we define the *likelihood* $L$ of the observed data as a function of $\mathcal{G}$, $\vec{C}$ and the parameters $\theta_{\mathcal{G}}$ as follows:

$$L(\mathcal{G}, \theta_{\mathcal{G}}, \vec{C}) = p(\vec{C}|\mathcal{G}^h, \theta_{\mathcal{G}})$$

where $\mathcal{G}^h$ is the hypothesis that the data was generated by a distribution that can be factored according to $\mathcal{G}$.[2]

---

[1] Researchers typically make this assumption for computational efficiency. Methods exist for scoring structures when there is missing data.

[2] This is an *acausal* interpretation of a network structure. Heckerman et al. (1994, 1995) investigate a causal interpretation of a network structure as well.

It follows by the definition of $\mathcal{G}^h$ that the hypotheses corresponding to equivalent structures are identical. We call this property *hypothesis equivalence*.

The *maximum likelihood metric* of a structure $\mathcal{G}$ is defined as

$$M_{ML}(\mathcal{G}, \vec{C}) = \max_{\theta_{\mathcal{G}}} L(\mathcal{G}, \theta_{\mathcal{G}}, \vec{C})$$

It follows almost immediately from the definition of equivalent structures that the maximum likelihood metric is score equivalent. Maximum likelihood is not very useful as a scoring metric by itself because any complete network structure will always get the highest possible score. Many of the metrics we are about to discuss, however, are defined as the sum of $M_{ML}$ and a penalty term.

The first scoring metric we consider, introduced by Akaike (1974), is called the A information criterion (AIC). In the context of scoring Bayesian networks, $M_{AIC}$ can be expressed as follows:

$$M_{AIC}(\mathcal{G}, \vec{C}) = \log M_{ML}(\mathcal{G}, \vec{C}) - Dim(\mathcal{G}) \quad (3)$$

**Theorem 4** $M_{AIC}$ *is score equivalent.*

**Proof:** Follows immediately from Theorem 3 and the fact that $M_{ML}$ is score equivalent. $\square$

Another scoring metric introduced by Schwarz (1978) is the Bayesian information criterion (BIC). This metric is defined as

$$M_{BIC}(\mathcal{G}, \vec{C}) = \log M_{ML}(\mathcal{G}, \vec{C}) - \frac{1}{2} Dim(\mathcal{G}) \log N$$

where $N$ is the number of cases in $\vec{C}$.

**Theorem 5** $M_{BIC}$ *is score equivalent.*

The proof of Theorem 5 is identical to the proof of Theorem 4.

Rissanen (1986) presents two scoring metrics using the principle of minimum description length (MDL). One of these metrics, originally presented in Rissanen (1978), has recently received some attention in the literature. A version of this metric explored by Bouckaert (1993b) is

$$\begin{aligned} M_{MDL1}(\mathcal{G}, \vec{C}) &= \log p(\mathcal{G}^h) - N \cdot H(\mathcal{G}, \vec{C}) \\ &\quad - \frac{1}{2} Dim(\mathcal{G}) \log N \end{aligned} \quad (4)$$

where $p(\mathcal{G}^h)$ is the prior probability of hypothesis $\mathcal{G}^h$, and $H(\mathcal{G}, \vec{C})$ is the entropy of the distribution resulting from parameterizing $\mathcal{G}$ with the appropriate fractions in the data. It can be shown that $-N \cdot H(\mathcal{G}, \vec{C})$ is identical to the log of the maximum likelihood metric. Therefore equation 4 reduces to

$$M_{MDL1}(\mathcal{G}, \vec{C}) = \log p(\mathcal{G}^h) + M_{BIC}(\mathcal{G}, \vec{C})$$

**Theorem 6** $M_{MDL1}$ *is score equivalent.*

**Proof:** Follows immediately from hypothesis equivalence and Theorem 5. □

Theorem 6 was also proven by Bouckaert (1993a). Another version of the MDL metric presented by Lam and Bacchus (1993) can be written as

$$M_{MDL2}(\mathcal{G}, \vec{C}) = -N \cdot H(\mathcal{G}, \vec{C}) - |E_{\mathcal{G}}| \log N - c \cdot Dim(\mathcal{G})$$

where $c$ is a constant that represents the number of bits needed to store a numerical value to some specified precision. We can again eliminate the entropy term to obtain

$$M_{MDL2}(\mathcal{G}, \vec{C}) = \log M_{ML}(\mathcal{G}, \vec{C}) - |E_{\mathcal{G}}| \log N - c \cdot Dim(\mathcal{G})$$

**Theorem 7** $M_{MDL2}$ is score equivalent.

**Proof:** The first term is score equivalent by definition of equivalent structures. The second term is score equivalent by Theorem 1. The third term is score equivalent by Theorem 3. □

The last metric we consider is a Bayesian metric discussed by Heckerman et al. (1994, 1995) known as the BDe metric. A Bayesian metric is any metric that expresses the relative posterior probability of the structure hypothesis, given the observed cases and prior knowledge. Specifically, for any Bayesian metric $M$ we have

$$M(\mathcal{G}, \vec{C}, \xi) = \log p(\mathcal{G}^h | \xi) + \log p(\vec{C} | \mathcal{G}^h, \xi) + c \quad (5)$$

The first term in Equation 5 is the prior probability of the structure hypothesis. The second term, which we call the *likelihood* of the data, is the posterior probability of the data given the structure hypothesis. The third term is an arbitrary constant.

Heckerman et al. (1994, 1995) derive a closed form expression for the likelihood of Equation 5 using some assumptions about prior densities over $\theta_{\mathcal{G}}$. Before we present the expression, we need the following notation. $q_i$ is the number of distinct instantiations of the parents of node $x_i$. $N_{ijk}$ is the number of cases in $\vec{C}$ for which $x_i = k$ and $\Pi_i$ in its $j$th configuration.

$$p(\vec{C} | \mathcal{G}^h, \xi) = \prod_{i=1}^{n} \prod_{j=1}^{q_i} \frac{\Gamma(N'_{ij})}{\Gamma(N'_{ij} + N_{ij})} \cdot \prod_{k=1}^{r_i} \frac{\Gamma(N'_{ijk} + N_{ijk})}{\Gamma(N'_{ijk})} \quad (6)$$

where $N'_{ijk} = p(x_i = k, \Pi_i = j | \xi)$, $N_{ij} = \sum_k N_{ijk}$ and $N'_{ij} = \sum_k N'_{ijk}$. $\Gamma$ is the *Gamma* function, which satisfies $\Gamma(x+1) = x\Gamma(x)$. In practice, researchers can use a prior network to determine both $p(x_i = k, \Pi_i = j | \xi)$ and a reasonable prior distribution $p(\mathcal{G}^h | \xi)$.

The BDe metric is defined to be the Bayesian metric of Equation 5 for which the likelihood term is computed using Equation 6. We say that a Bayesian metric is *likelihood equivalent* if $p(\vec{C} | \mathcal{G}^h, \xi)$ is score equivalent. Heckerman et al. (1994, 1995) use Theorem 2 to prove the following result.

**Theorem 8** $M_{BDe}$ is likelihood equivalent.

It follows by hypothesis equivalence that $M_{BDe}$ is score equivalent as well.[3]

## 3.3  NUMBER OF PARENTS

In this section we present yet another consequence of Theorem 2. This result was used by Chickering et al. (1995) to prove that the Bayesian approach to learning Bayesian networks from data is NP-hard.

**Theorem 9** Let $\mathcal{G}$ and $\mathcal{G}'$ be any pair of dags such that $\mathcal{G} \approx \mathcal{G}'$. If $\mathcal{G}$ has a node with exactly $k$ parents, then $\mathcal{G}'$ has a node with exactly $k$ parents.

**Proof:** From Theorem 2, we need only show that the theorem holds when $\mathcal{G}$ and $\mathcal{G}'$ differ by the orientation of a single covered edge. Let $x \rightarrow y$ be this edge in $\mathcal{G}$.

Because every node except for $x$ and $y$ have identical parents in $\mathcal{G}$ and $\mathcal{G}'$, the theorem holds trivially unless the only node in $\mathcal{G}$ that has $k$ parents is either $x$ or $y$. From the definition of a covered edge it follows that in $\mathcal{G}'$, $x$ has the same number of parents that $y$ has in $\mathcal{G}$ and $y$ has the same number of parents that $x$ has in $\mathcal{G}$. □

# 4  IDENTIFYING COMPELLED EDGES

In this section we first discuss the significance of compelled edge identification for learning networks from data and explore previous relevant work. Next we present an algorithm that identifies the set of all compelled edges in an equivalence class. Finally, we discuss an implementation of the algorithm that is asymptotically optimal on average.

As was mentioned in Section 1, identifying compelled edges is of particular importance for learning Bayesian networks from data because these edges can indicate causal influences. The assumptions needed to infer causation from the compelled edges are (1) if two variables are statistically dependent in every observable context, then one of the variables is a direct cause of the other, and (2) if two variables are statistically independent in some (possibly empty) observable context, then neither variable is a direct cause of the other. Note that the first assumption excludes the possibility that there is a hidden common cause of two variables. Spirtes et al. (1993) call Assumption 1 *causal sufficiency* and Assumption 2 *faithfulness*. Assumption 2 is called *stability* by Pearl and Verma (1991) . If an equivalence class is learned with certainty and the assumptions hold, then all the compelled edges denote causal influences.

---

[3]Theorem 8 was proven without the assumption of hypothesis equivalence. It follows that the result also applies to causal interpretations of structures for which hypothesis equivalence does not hold.

There are two distinct approaches that researchers use to learn Bayesian networks from data. The first approach, which we call the *metric approach*, uses a scoring metric to measure how well a particular structure fits an observed set of cases. A search algorithm is typically used to identify one or more structures that attain a high metric score. As we saw in Section 3.2, many of the metrics that researchers use have the property of score equivalence. It follows that when using a score equivalent metric, the metric approach to learning is in fact a method for identifying entire equivalence classes. We assume in this section that a score equivalent metric is being used in the metric approach.

In the second approach to learning, which we call the *independence* approach, an independence oracle is queried to identify the equivalence class that captures the independencies in the distribution from which the observed data was generated.

One distinction between the two learning approaches is the way equivalence classes are represented. Using the metric approach, an equivalence class is represented by any element in the class. We call this the *canonical element* representation scheme. In the independence approach, researchers typically use an acyclic partially directed graph (*pdag* for short) to represent the equivalence class.

From Theorem 1, the only edges that need be directed in a pdag representation to uniquely identify an equivalence class are those that participate in v-structures. If, in fact, these are the only directed edges in the pdag, we say that the graph is a *minimal pdag representation* of the equivalence class. If a pdag has the property that every directed edge corresponds to a compelled edge, and every undirected edge corresponds to a reversible edge for every dag in the equivalence class, then we say it is a *completed* pdag representation.

When using the independence approach to learning, researchers use a statistical test (such as chi-square) to approximate the independence oracle, and the learning algorithm builds a unique minimal pdag representation of the equivalence class. We refer the reader to Verma and Pearl (1992) or Spirtes et al. (1993) for the details of this procedure.

Previous work on compelled edge identification can be understood in the context of the independence approach to learning. After identifying a minimal pdag representation of an equivalence class, the learning algorithm searches for the remaining compelled edges that do not participate in v-structures by matching patterns of directed and undirected edges. When a match is found, one or more undirected edges in the pdag are directed and the process continues. Verma and Pearl (1992) present an algorithm of this type that is known to be sound, but not complete. That is, every directed edge in the final pdag is provably compelled, but not every undirected edge is provably reversible. More recently, both Meek (1995) and Anderson et al. (1995) have derived sound and complete algorithms for constructing a completed pdag representation.

The algorithm we present in this section takes a dag as input, and labels every edge in the dag as either compelled or reversible. We show that the algorithm is correct and discuss an implementation that is asymptotically optimal in the average case. Our algorithm is also applicable to identifying compelled edges when the equivalence class is represented with a pdag. Dor and Tarsi (1992) present a polynomial-time algorithm that takes as input a (possibly minimal) pdag representation of an equivalence class, and returns a canonical element representation. Unfortunately, the algorithm has a worst-case time complexity that is worse than that of of our identification algorithm. Nonetheless, if the equivalence class is represented as a minimal pdag, we get a significant improvement in asymptotic behavior over the previous algorithms by changing representations and using our algorithm.

When the independence approach to learning is used, an equivalence class is determined with certainty. Because the data is finite, however, we have uncertainty about the learned equivalence class as a result of the approximation of the independence oracle. One advantage to using a Bayesian scoring metric instead of other metrics or the independence approach is that the uncertainty about any equivalence class is explicitly represented in the score of that class. Consequently, we can express our belief in the statement $s =$ "x is a direct cause of y" by summing over all (non-equivalent) structures:

$$
\begin{aligned}
p(s|\vec{C}, \xi) &= \sum_{\mathcal{G}} p(s|\mathcal{G}^h, \vec{C}, \xi) \cdot p(\mathcal{G}^h|\vec{C}, \xi) \\
&= \sum_{\mathcal{G}} p(s|\mathcal{G}^h, \xi) \cdot p(\mathcal{G}^h|\vec{C}, \xi) \qquad (7)
\end{aligned}
$$

In practice, it is impossible to sum over all possible equivalence classes. Therefore we attempt to find a small subset of structure hypotheses that account for a large fraction of the posterior probability of the hypotheses. Chickering (1995) suggests a set of search operators that can be used to efficiently search for such a subset.

For those hypotheses in which $x \rightarrow y$ is compelled, the corresponding probability $p(s|\mathcal{G}^h, \xi)$ term will be unity. If $y \rightarrow x$ is compelled or if $x$ and $y$ are not adjacent, then the term will be zero. If the edge between $x$ and $y$ is reversible, then $p(s|\mathcal{G}^h, \xi)$ can take any value from zero to one and must be assessed directly.

Heckerman et al. (1994, 1995) discuss a causal interpretation for the hypothesis $\mathcal{G}^h$. The hypothesis not only asserts that the distribution that generated $\vec{C}$ can be factored according to $\mathcal{G}$, but that each node in the graph is a direct cause of its children. Using this interpretation, the term $p(s|\mathcal{G}^h, \xi)$ is either one or zero, depending on whether $x \rightarrow y$ is in $\mathcal{G}$ or not. When structures are interpreted causally, we no longer have the property of hypothesis equivalence. Nonetheless, we can sometimes still use entire equivalence classes to calculate Equation 7. For example, if the prior probability distribution over structure hypotheses is uniform, then a likelihood equivalent learning metric is score equivalent. In this case, we can still take the sum in Equation 7 over non-equivalent structures and weight each term by the number of dags in the equivalence class that contain the edge $x \rightarrow y$. We are currently investigating techniques for efficiently determining such a weighting term.

In Section 4.1 we present the algorithm and prove that it correctly classifies all of the edges in a dag. In Section 4.2 we discuss asymptotic running time behavior of various implementations.

## 4.1  THE ALGORITHM

The first step of the algorithm is to define a total ordering over the edges in the given dag. For simplicity, we present this step as a separate procedure listed below. To avoid confusion between ordered nodes and ordered edges, we have capitalized "node" and "edge" below.

**Algorithm Order-Edges($\mathcal{G}$)**
**Input:** dag $\mathcal{G}$
**Output:** dag $\mathcal{G}$ with labelled total order on edges

1. Perform a topological sort on the NODES in $\mathcal{G}$
2. Set $i = 0$
3. While there are unordered EDGES in $\mathcal{G}$
4.    Let $y$ be the lowest ordered NODE that has an unordered EDGE incident into it
5.    Let $x$ be the highest ordered NODE for which $x \rightarrow y$ is not ordered
6.    Label $x \rightarrow y$ with order $i$
7.    $i = i + 1$

The algorithm to find the compelled edges is as follows.

**Algorithm Find-Compelled($\mathcal{G}$)**
**Input:** dag $\mathcal{G}$
**Output:** dag $\mathcal{G}$ with each edge labelled either "compelled" or "reversible"

1. Order the edges in $\mathcal{G}$ using **Algorithm Order-Edges**
2. Label every edge in $\mathcal{G}$ as "unknown"
3. **While** there are edges labelled "unknown" in $\mathcal{G}$
4.    Let $x \rightarrow y$ be the lowest ordered edge that is labelled "unknown"
5.    For every edge $w \rightarrow x$ labelled "compelled"
6.       **If** $w$ is not a parent of $y$, then label $x \rightarrow y$ and every edge incident into $y$ with "compelled" and goto 3
7.       **Else** label $w \rightarrow y$ with "compelled"
8.    **If** there exists an edge $z \rightarrow y$ such that $z \neq x$ and $z$ is not a parent of $x$, then label $x \rightarrow y$ and all "unknown" edges incident into $y$ with "compelled"

9.    **Else** label $x \rightarrow y$ and all "unknown" edges incident into $y$ with "reversible"

Before proving the correctness of the algorithm, we need a few intermediate results. The proofs of the first two results, which are given in the Appendix, make extensive use of Theorem 2.

**Lemma 3** *Let $\mathcal{G}$ be any dag and let $x$, $y$ and $z$ be any three nodes that are all adjacent in $\mathcal{G}$. If any two of the connecting edges are reversible, then the third one is also.*

**Lemma 4** *Let $\mathcal{G}$ be any dag, and let $x \rightarrow y$ be any edge in $\mathcal{G}$ such that $\Pi_y \subseteq \Pi_x \cup x$. The edge $x \rightarrow y$ is reversible if and only if for every edge $w \rightarrow x$ such that $w$ and $y$ are not adjacent, $w \rightarrow x$ is reversible.*

In addition to the above two lemmas, we find the following three simple results useful for proving the correctness of our algorithm.

**Lemma 5** *When $x \rightarrow y$ is chosen in Step 4 of the algorithm, every edge incident into node $y$ is labelled "unknown".*

**Proof:** Follows by noting that after any iteration of the while loop, every edge incident into $y$ gets labelled with either "reversible" or "compelled". $\square$

**Lemma 6** *Let $x \rightarrow y$ be the edge chosen in Step 4 of the algorithm. Any parent of $y$ that is adjacent to $x$ is a parent of $x$.*

**Proof:** Let $z$ be any parent of $y$ that is adjacent to $x$. By Lemma 5 we know $z \rightarrow y$ is labelled "unknown". If $x \rightarrow z$, then $z \rightarrow y$ has a lower order than $x \rightarrow y$ (see **Algorithm Order-Edges**) and would have been chosen in Step 4. $\square$

**Lemma 7** *Let $x \rightarrow y$ be the edge chosen in Step 4 of the algorithm. If $x \rightarrow y$ is compelled, then every edge incident into $y$ is compelled.*

**Proof:** Let $z \rightarrow y$ be any edge incident into $y$. If $z$ and $x$ are not adjacent, then $z \rightarrow y$ must be compelled because it participates in a v-structure. If $z$ and $x$ are adjacent, then from Lemma 6 we know the edge is oriented as $z \rightarrow x$. If $z \rightarrow x$ is compelled, then there is a directed path from $z$ to $y$ in every graph equivalent to $\mathcal{G}$ and hence reversing $z \rightarrow y$ will always create a cycle. If $z \rightarrow x$ is in $R_\mathcal{G}$, then by Lemma 3 $z \rightarrow y$ is compelled. $\square$

Now we prove the correctness of our algorithm.

**Theorem 10** *The edge labels resulting from the algorithm are correct.*

**Proof:** We prove that the labellings are correct by induction on the number of iterations through the while loop.

When the first iteration of the while loop begins, all edges are labelled "unknown". Thus for the edge $x \rightarrow y$ chosen in Step 4, there can be no edge $w \rightarrow x$ tested for in Step 5, lest this would be the first edge instead of $x \rightarrow y$. Therefore $\Pi_x = \emptyset$ and the algorithm drops immediately to Step 8. There can be no edge $z \rightarrow y$ such that $z$ and $x$ are adjacent because by Lemma 6, we know that any edge between $z$ and $x$ is oriented as $z \rightarrow x$ which implies that $x \rightarrow y$ would not have been the first edge chosen. Therefore if any edge $z \rightarrow y$ is incident into $y$, that edge is part of a v-structure with $x \rightarrow y$ and therefore both $z \rightarrow y$ and $x \rightarrow y$ are compelled. Furthermore, by Lemma 7, all edges incident into $y$ are compelled, so all labelling done in Step 8 is correct. If Step 9 is reached, no edge $z \rightarrow y$ exists and it follows that $\Pi_y = x$. Because $\Pi_x = \emptyset$, $x \rightarrow y$ is reversible by Lemma 4, and the labelling done at Step 9 is correct.

Assume all labelling is correct for the first $k - 1$ iterations through the while loop of Step 3. Consider the edge $x \rightarrow y$ chosen in Step 4 on the $kth$ iteration of the while loop.

From Step 6, if there is a compelled edge $w \rightarrow x$ such that $w$ is not a parent of $y$ then $w$ and $y$ are not adjacent, lest $\mathcal{G}$ contains a directed cycle. It follows that $x \rightarrow y$ must be compelled, lest there would exist a dag in the same equivalence class as $\mathcal{G}$ with the extra v-structure $(w, x, y)$. By Lemma 7, it follows that all edges incident into $y$ are compelled and therefore all labelling done in Step 6 is correct. If there is a compelled edge $w \rightarrow x$ such that $w$ and $y$ are adjacent, then the edge between $w$ and $y$ must be oriented as $w \rightarrow y$ lest $\mathcal{G}$ would contain a directed cycle. Furthermore, we deduce this edge is compelled by the following argument: if $x \rightarrow y$ is compelled, then there is a directed path from $w$ to $y$ in every dag equivalent to $\mathcal{G}$ and hence reversing $w \rightarrow y$ will always create a cycle; if $x \rightarrow y$ is reversible, then $w \rightarrow y$ is compelled by Lemma 3. Thus all labelling done in Step 7 is correct.

From Step 8, if there exists a parent $z$ of $y$ that is not a parent of $x$, then by Lemma 6, $z$ is not adjacent to $x$ which implies that $x \rightarrow y$ participates in a v-structure and is therefore compelled. Furthermore, by Lemma 7, all edges incident into $y$ are compelled and hence all labelling done in Step 8 is correct.

If Step 9 is reached, every parent of $y$ (with the exception of $x$) is a parent of $x$. That is, $\Pi_y \subseteq \Pi_x \cup x$. Furthermore, because Step 9 is reached only if Step 6 always fails, every edge $w \rightarrow x$ for which $w$ and $y$ are not adjacent must be reversible. Consequently, we conclude from Lemma 4 that $x \rightarrow y$ is reversible.

Now consider any edge $z \rightarrow y$ incident into $y$ that is labelled with "unknown". It must be the case that $z \rightarrow x$ is reversible, lest $z \rightarrow y$ would have been labelled "compelled" in Step 7. Thus we conclude from Lemma 3 that $z \rightarrow y$ is reversible and hence all labelling done in Step 9 is correct. □

## 4.2  COMPLEXITY ANALYSIS

In this section we investigate the asymptotic time behavior of various implementations of **Algorithm Find-Compelled** presented in Section 4.1. Because the algorithm labels every edge in the dag $\mathcal{G}$, the best that any implementation can do is $O(|E_{\mathcal{G}}|)$.

We first investigate an implementation of **Algorithm Order-Edges** that runs in time $O(|E_{\mathcal{G}}|)$. We assume that $\mathcal{G}$ is represented using the adjacency-list representation. It is well known that a topological sort can be performed in time $O(|E_{\mathcal{G}}|)$ using a depth-first search. Once the nodes in the dag have been ordered, we would like to sort the parents of each node in *descending* order. Once we have accomplished this, sorting the edges becomes trivial: step through each node in ascending sort order, and for each node, list all the incident edges by stepping through the sorted parent list.

One simple way to sort the parent pointers is as follows. Extend the representation to include child-pointers for each node. This will take time $O(|E_{\mathcal{G}}|)$. Now step through each node in ascending order, and for each child of the current node, insert the current node at the *front* of the parent list. When the algorithm completes (in time $O(|E_{\mathcal{G}}|)$) the parent pointers will be sorted in descending order.

From the above discussion, we see that Step 1 of **Algorithm Find-Compelled** can be completed in time $O(|E_{\mathcal{G}}|)$. Assume that with each node, we store separate lists of "compelled", "reversible", and "unknown" incident edges (i.e. parents), so that these can be efficiently accessed in the algorithm. We now consider the inside of the while loop. For each edge that we consider in Step 5, we necessarily label at least one "unknown" edge in either Step 6 or Step 7. Thus neither Step 5, Step 6 nor Step 7 can ever be executed more than $|E_{\mathcal{G}}|$ times. Furthermore, for every edge considered in Step 8, that edge is "unknown" (See Lemma 7) and will get labelled in the current iteration of the while loop. Thus every operation within the while loop is executed no more than $|E_{\mathcal{G}}|$ times.

We note that it is possible to get an amortized constant time adjacency test in Step 6, but do not want to worry the reader with the details. Unfortunately, this is not the case for the adjacency test in Step 8. Because all labellings can be done in constant time, it follows that the time complexity of the entire algorithm is dominated by the $O(|E_{\mathcal{G}}|)$ executions of Step 8. If we use a hash table to store the parents of each node, we can complete the adjacency test in constant time on average. The resulting implementation of the algorithm will take time $O(|E_{\mathcal{G}}|)$ on average, which is asymptotically optimal.

A problem with the hash table implementation is that in the worst case, each adjacency test can take $O(|U|)$, resulting in a worst case $O(|U| \cdot |E_{\mathcal{G}}|)$ algorithm. If instead we test for adjacency by performing a binary search over the parents of a node, each test can be

completed in time $O(\log|U|)$, and the resulting algorithm takes time $O(|E_{\mathcal{G}}|\log|U|)$ in the worst case.

If $\mathcal{G}$ is represented with an adjacency matrix, then testing for adjacency will always be a constant time operation, and therefore **Algorithm Find-Compelled** takes time $O(|E_{\mathcal{G}}|)$ in the worst case. Building the adjacency matrix, however, takes time $O(|U|^2)$ and will therefore dominate the time to complete the algorithm. For dense graphs, the use of an adjacency matrix is a good solution.

If an equivalence class is represented using a minimal pdag, we can construct a completed pdag using a combination of our algorithm and the algorithm presented by Dor and Tarsi (1992) which returns a canonical element given a minimal pdag. First we obtain the canonical element, which has been shown to take time $O(|U| \cdot |E_{\mathcal{G}}|)$. Next we run **Algorithm Find-Compelled** to determine all the compelled edges. Finally we direct every undirected edge in the original pdag that corresponds to a compelled edge. The running time of the combined algorithm is dominated by the time to retrieve the canonical element and is therefore $O(|U| \cdot |E_{\mathcal{G}}|)$.

## Acknowledgments

I would like to thank David Galles, Dan Geiger, Rich Korf, David Madigan, Chris Meek, Judea Pearl, and anonymous reviewers for useful suggestions. I owe special thanks to David Heckerman, whose help and encouragement made this work possible. This work was supported by NSF Grant No. IRI-9119825, and a grant from Rockwell International.

## APPENDIX: PROOF OF LEMMAS 3 AND 4

In this appendix, we prove Lemma 3 and Lemma 4, using numerous intermediate results.

For any pair of dags $\mathcal{G}$ and $\mathcal{G}'$ that share the same skeleton, we use $\delta_i(\mathcal{G}, \mathcal{G}')$ to be the set of edges incident into node $x_i$ in $\mathcal{G}$ that have opposite orientation in $\mathcal{G}'$. Note that $\Delta(\mathcal{G}, \mathcal{G}') = \cup_i \delta_i(\mathcal{G}, \mathcal{G}')$

For many of the lemmas to follow, we consider the ordered sequence of intermediate dags — and the ordered sequence of edge reversals that created the intermediate dags — in a transformation from some dag $\mathcal{G}$ to another dag $\mathcal{G}' \approx \mathcal{G}$ using the procedure as described in the proof of Theorem 2. To make our discussion clear, we provide the following detailed description of the algorithm from the proof of Theorem 2 that has been modified to build the desired sequences, as opposed to actually modifying the dag $\mathcal{G}$.

**Build-Sequences($\mathcal{G}, \mathcal{G}'$)**

1. Set $\mathcal{G}_0 = \mathcal{G}$ and Set $i = 0$
2. While $\mathcal{G}_i \neq \mathcal{G}'$
3.      Let $e_i =$ **Find-Edge($\mathcal{G}_i, \mathcal{G}'$)**
4.      Set $\mathcal{G}_{i+1}$ to be the result of reversing $e_i$ in $\mathcal{G}_i$
5.      Increment $i$ by one

We use $\mathcal{D}(\mathcal{G}, \mathcal{G}') = \{\mathcal{G}_0, \ldots, \mathcal{G}_{|\Delta(\mathcal{G},\mathcal{G}')|}\}$ for the ordered sequence of dags constructed from the above algorithm. Similarly, we use $\mathcal{E}(\mathcal{G}, \mathcal{G}') = \{e_0, \ldots, e_{|\Delta(\mathcal{G},\mathcal{G}')|-1}\}$ for the ordered sequence of edges reversed in the above algorithm. Note that given $\mathcal{G}_i \in \mathcal{D}(\mathcal{G}, \mathcal{G}')$ we can construct $\mathcal{G}_{i+1} \in \mathcal{D}(\mathcal{G}, \mathcal{G}')$ by reversing $e_i \in \mathcal{E}(\mathcal{G}, \mathcal{G}')$ in $\mathcal{G}_i$.

The sequences $\mathcal{D}(\mathcal{G}, \mathcal{G}')$ and $\mathcal{E}(\mathcal{G}, \mathcal{G}')$ depend not only on $\mathcal{G}$ and $\mathcal{G}'$, but on the specific topological sort performed in Step 1 of each call to **Algorithm Find-Edge**. Because the topological sort may not be unique, it seems that our definition of these two sets is ambiguous. As we shall see, however, the topology of $\mathcal{G}$ constrains the sequences enough for our current definition to be useful.

**Lemma 8** *Let $\mathcal{G}_i$ and $\mathcal{G}_{i+1}$ be any pair of dags in the sequence $\mathcal{D}(\mathcal{G}, \mathcal{G}')$. Let $e_i = x_t \rightarrow x_h$ be the edge by*

*which $\mathcal{G}_i$ and $\mathcal{G}_{i+1}$ differ. Then the following conditions hold:*

    *1. $\delta_h(\mathcal{G}_{i+1}, \mathcal{G}') = \delta_h(\mathcal{G}_i, \mathcal{G}') \setminus e_i$*

    *2. $\delta_j(\mathcal{G}_{i+1}, \mathcal{G}') = \delta_j(\mathcal{G}_i, \mathcal{G}')$ for all $j \neq h$*

**Proof:** Condition 1 follows trivially because the only difference between $\mathcal{G}_i$ and $\mathcal{G}_{i+1}$ is the orientation of $e_i$. The only parent sets that have changed as a result of the reversal are $\Pi_t$ and $\Pi_h$, and hence Condition 2 holds when $j$ is neither $t$ nor $h$. Because $x_t \rightarrow x_h$ is in $\delta_h(\mathcal{G}_i, \mathcal{G}')$, the reversed edge $x_h \rightarrow x_t$ cannot be in $\delta_t(\mathcal{G}_{i+1}, \mathcal{G}')$ and hence Condition 2 holds as stated. $\square$

**Corollary 1** *For any node $x_i$, $\delta_i(\mathcal{G}_k, \mathcal{G}') \subseteq \delta_i(\mathcal{G}_j, \mathcal{G}')$ if $j < k$.*

**Proof:** Follows immediately from Lemma 8. $\square$

Corollary 1 shows formally that as we progress along the sequence $\mathcal{D}(\mathcal{G}, \mathcal{G}')$, the edges incident into a particular node that have different orientations in $\mathcal{G}'$ is a strictly decreasing set.

**Lemma 9** *Let $x_i \rightarrow x_j$ be the edge returned by a call to Algorithm Find-Edge($\mathcal{G}, \mathcal{G}'$). Then $\delta_k(\mathcal{G}, \mathcal{G}')$ is empty for every $x_k$ that is an ancestor of $x_j$.*

**Proof:** Suppose there exists a node $x_k$ that is an ancestor of $x_j$ for which $\delta_k(\mathcal{G}, \mathcal{G}')$ is not empty. In any topological sort consistent with $\mathcal{G}$, $x_k$ must precede $x_j$ and hence $x_k$ would have been chosen instead of $x_j$ in Step 2 of **Algorithm Find-Edge($\mathcal{G}, \mathcal{G}'$)**. $\square$

**Lemma 10** *Let $\mathcal{G}$ be any dag, and let $x_i$ and $x_j$ be any pair of nodes such that there is a directed path from $x_i$ to $x_j$ in $\mathcal{G}$. Let $\mathcal{G}'$ be any dag equivalent to $\mathcal{G}$. For any pair of edges $e \in \delta_i(\mathcal{G}, \mathcal{G}')$ and $f \in \delta_j(\mathcal{G}, \mathcal{G}')$, e comes before f in $\mathcal{E}(\mathcal{G}, \mathcal{G}')$.*

**Proof:** (See Figure 1) Without loss of generality, let $x_j$ be the *first* descendant of $x_i$ that has an incident edge $f$ reversed. Let $\mathcal{G}_k$ be the graph in which $f$ is reversed, or equivalently, let $k$ be the index such that $e_k = f$.

Assume the lemma does not hold, and hence $f$ is reversed before $e$. Because $x_j$ is the first descendant to have an incident edge reversed, it follows that any directed path from $x_i$ to $x_j$ in $\mathcal{G}$ must still exist in $\mathcal{G}_k$, and therefore $x_i$ is an ancestor of $x_j$ in $\mathcal{G}_k$. Because $f$ is the next edge to be reversed, it follows from Lemma 9 that $\delta_i(\mathcal{G}_k, \mathcal{G}')$ is empty. But by assumption $e$ has not yet been reversed in $\mathcal{G}_k$, and hence $e \subseteq \delta_i(\mathcal{G}_k, \mathcal{G}')$, yielding a contradiction. $\square$

**Lemma 11** *Let $\mathcal{G}$ be any dag, and let $R_y$ be any set of edges incident into node $y$ such that $R_y \subseteq R_{\mathcal{G}}$. Then there exists a dag $\mathcal{G}' \approx \mathcal{G}$ for which all edges in $R_y$ are simultaneously reversed.*

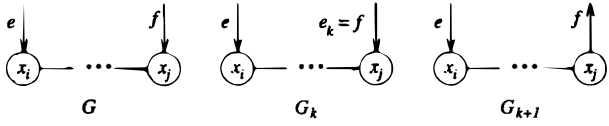

Figure 1: Relevant dags for the proof of Lemma 10

**Proof:** (See Figure 2) Proof by induction on $|R_y|$. For $|R_y| = 1$, the lemma holds trivially by definition of $R_{\mathcal{G}}$.

Assume the lemma holds for $|R_y| = k - 1$. We now show the lemma holds for $|R_y| = k$. By the induction hypothesis, there must exist a dag $\mathcal{H} \approx \mathcal{G}$ in which $k-1$ of the edges from $R_y$ are reversed. Let $R_{\mathcal{H}}$ be the corresponding set of reversed edges in $\mathcal{H}$, and let $e$ be the edge in $R_y$ that is *not* reversed in $\mathcal{H}$. By definition of $R_{\mathcal{G}}$, we know there exists some dag $\mathcal{H}' \approx \mathcal{G}$ for which the edge $e$ is reversed.

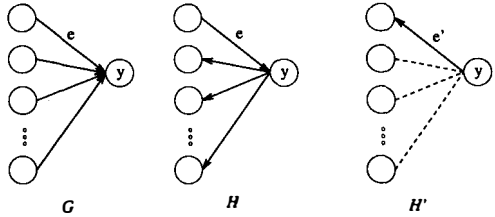

Figure 2: Relevant dags for the proof of Lemma 11. In dag $\mathcal{H}$, every edge from $R_y$ except $e$ has been reversed. In dag $\mathcal{H}'$, $e$ has been reversed.

Because all of the edges in $R_{\mathcal{H}}$ are incident into descendants of $y$ in $\mathcal{H}$, it follows from Lemma 10 that $e$ comes before any edge from $R_{\mathcal{H}}$ in the sequence $\mathcal{E}(\mathcal{H}, \mathcal{H}')$. This implies that the graph from $\mathcal{D}(\mathcal{H}, \mathcal{H}')$ that results from reversing $e$ satisfies the stated requirements for $\mathcal{G}'$. □

For any dag $\mathcal{G}$, we use $C_X(\mathcal{G})$ to denote the subgraph of $\mathcal{G}$ induced by the nodes in set $X$. A *clique* in a directed graph $\mathcal{G}$ is a subgraph $C_X(\mathcal{G})$ such that for every pair of nodes $x_i$ and $x_j$ in $X$, either the edge $x_i \rightarrow x_j$ or the edge $x_j \rightarrow x_i$ is in $\mathcal{G}$. A *covered clique* in a directed graph $\mathcal{G}$ is a clique $C_X(\mathcal{G})$ such that for any node $z \notin X$ that is a parent of some node in $X$, $z$ is a parent of every node in $X$. Note that a covered edge is a covered clique with two nodes.

**Lemma 12** *Let $\mathcal{G}$ be any dag containing a covered clique $C_X(\mathcal{G})$. No edge connecting a pair of nodes in $X$ can participate in a v-structure in $\mathcal{G}$.*

**Proof:** Suppose $x_i \rightarrow x_j$ connects two nodes in $X$ and participates in a v-structure. This implies there is a parent of $x_j$ that is not adjacent to $x_i$ — and hence not a parent of $x_i$ — contradicting the fact that the nodes in $X$ form a covered clique. □

The following lemma is a generalization of Lemma 1.

**Lemma 13** *Let $\mathcal{G}$ be any dag containing a covered clique $C_X(\mathcal{G})$, and let $\tau_X$ be any total ordering over the nodes in $X$. Let $\mathcal{G}'$ be the graph identical to $\mathcal{G}$, except the edges in $C_X(\mathcal{G}')$ are oriented to be consistent with $\tau_X$. Then $\mathcal{G}'$ is a dag that is equivalent to $\mathcal{G}$.*

**Proof:** Clearly $\mathcal{G}'$ and $\mathcal{G}$ have the same skeleton. Suppose there exists a v-structure in $\mathcal{G}$ that is not in $\mathcal{G}'$. This v-structure must include an edge from $C_X(\mathcal{G})$ because these are the only edges by which $\mathcal{G}$ can differ from $\mathcal{G}'$. But by Lemma 12, no such v-structure can exist.

Suppose there exists a v-structure in $\mathcal{G}'$ that is not in $\mathcal{G}$. Because only edges contained within $C_X(\mathcal{G})$ have been reversed in $\mathcal{G}$, it follows that $C_X(\mathcal{G}')$ must be a covered clique in $\mathcal{G}'$, and again by Lemma 12 we conclude that no such v-structure can exist.

Suppose $\mathcal{G}'$ contains a cycle (see Figure 3). Because $\mathcal{G}$ is acyclic, any cycle in $\mathcal{G}'$ must pass through an edge in $C_X(\mathcal{G}')$. Because these edges are consistent with the total ordering $\tau_X$, no cycle can be completely contained within $C_X(\mathcal{G}')$. This implies there exists a pair of nodes $x_i$ and $x_j$ in $X$ such that there is a directed path from $x_i$ to $x_j$ along the cycle in $\mathcal{G}'$ consisting of no edges from $C_X(\mathcal{G})$. Let $z$ be the last node in such a path. By definition of a covered clique, $z$ must be a parent of $x_i$ in $\mathcal{G}$, and therefore $\mathcal{G}$ contains a cycle. □

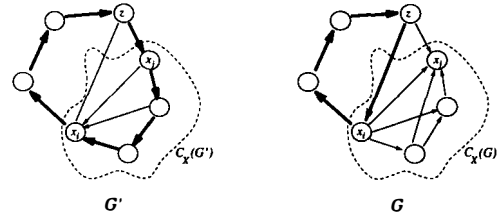

Figure 3: Relevant dags for the proof of Lemma 13. The dashed line surrounds the covered clique in both graphs.

**Lemma 14** *Let $\mathcal{G}$ be any dag, and let $R_y = \{x_1 \rightarrow y, \ldots, x_k \rightarrow y\}$ be any subset of edges incident into node $y$ such that $R_y \subseteq R_{\mathcal{G}}$, and let $X = \{x_1, \ldots, x_k\}$ be the set of tails of these edges. There exists a dag $\mathcal{G}' \approx \mathcal{G}$ for which $C_{X \cup \{y\}}(\mathcal{G}')$ is a covered clique in $\mathcal{G}'$.*

**Proof:** By Lemma 11, we know there exists a dag $\mathcal{H} \approx \mathcal{G}$ such that $R_y \subseteq \delta_y(\mathcal{G}, \mathcal{H})$. Consider the sequence $\mathcal{D}(\mathcal{G}, \mathcal{H})$, and let $\mathcal{G}'$ be the dag resulting immediately after the last edge from $R_y$ is reversed. We show that $C_{X \cup \{y\}}(\mathcal{G}')$ is a covered clique in $\mathcal{G}'$.

First we show that $C_{X \cup \{y\}}(\mathcal{G}')$ is a clique. By definition of $X$, $y$ is adjacent to every node in $X$. Thus if the subgraph $C_{X \cup \{y\}}(\mathcal{G}')$ is not a clique, then there must exist some pair $\{x_i, x_j\}$ from $X$ that are not adjacent. But this implies that $(x_i, y, x_j)$ is a v-structure in $\mathcal{G}$, contradicting the fact that both $x_i \rightarrow y$ and $x_j \rightarrow y$ are members of $R_{\mathcal{G}}$.

To complete the proof, we must show that for any node $z$ that is not in $X \cup \{y\}$, if $z$ is a parent of any node in $X \cup \{y\}$, then $z$ is a parent of every node in $X \cup \{y\}$. We break this task into two parts: we show that in $\mathcal{G}'$, (1) if $z$ is a parent of $y$ then $z$ is a parent of every node in $X$ and (2) if $z$ is a parent of any node in $X$, then $z$ is a parent of $y$.

**(1)** If $z$ is a parent of $y$ in $\mathcal{G}'$ then $z$ is a parent of every node in $X$ in $\mathcal{G}'$.

Assume $z$ is a parent of $y$.

We first show that $z$ is also a parent of $y$ in the original dag $\mathcal{G}$. Suppose this is not the case, and $\mathcal{G}$ contains the edge $y \rightarrow z$. By Lemma 10 all edges from $R_y$ will be reversed before this edge. It follows by definition of $\mathcal{G}'$, however, that the last edge reversed was incident into $y$ and hence $y \rightarrow z$ must exist in $\mathcal{G}'$, contradicting the assumption that $z$ is a parent of $y$ in $\mathcal{G}'$.

Now we prove Part (1) by showing that $z$ is a parent of every node $x_i \in X$. For any node $x_i \in X$, $z$ must be adjacent to $x_i$ else the v-structure $(x_i, y, z)$ exists in $\mathcal{G}$ and not $\mathcal{G}'$. The edge must be oriented as $z \rightarrow x_i$ in $\mathcal{G}'$, else there would be the directed cycle $x_i \rightarrow z \rightarrow y \rightarrow x_i$ in $\mathcal{G}'$.

**(2)** If $z$ is a parent of any node in $X$ in $\mathcal{G}'$, then $z$ is a parent of $y$ in $\mathcal{G}'$.

Assume $z$ is a parent of some node $x_i \in X$ in $\mathcal{G}'$.

The node $z$ must be adjacent to $y$, lest the v-structure $(z, x_i, y)$ exists in $\mathcal{G}'$ and not in $\mathcal{G}$.

We now consider the two possible orientations for the edge between $z$ and $y$ in the original graph $\mathcal{G}$. If the edge $y \rightarrow z$ is in $\mathcal{G}$, then $\mathcal{G}$ must also contain $x_i \rightarrow z$, lest there would be a directed cycle in $\mathcal{G}$. Because $z$ is a descendent of $y$ in $\mathcal{G}$, we know from Lemma 10 that all the edges from $R_y$ will be reversed before $x_i \rightarrow z$. It follows by definition of $\mathcal{G}'$, however, that the last edge reversed was incident into $y$ and hence $x_i \rightarrow z$ must also exist in $\mathcal{G}'$, contradicting the assumption that $z$ is a parent of $x_i$ in $\mathcal{G}'$.

It follows from the above argument that $z \rightarrow y$ must exist in the original dag $\mathcal{G}$. Now, if $y \rightarrow z$ is in $\mathcal{G}'$, then $z \rightarrow y \in R_y$, contradicting the fact that $z \notin X$. Consequently, $z$ must be a parent of $y$ in $\mathcal{G}'$. □

**Corollary 2** *Let $\mathcal{G}$ be any dag, and let $R_y = \{x_1 \rightarrow y, \ldots, x_k \rightarrow y\}$ be any set of edges incident into node $y$ such that $R_y \subseteq R_{\mathcal{G}}$. Let $X = \{x_1, \ldots, x_k\}$ be the set of tails of these edges. Every edge in $C_{X \cup \{y\}}(\mathcal{G})$ is in $R_{\mathcal{G}}$*

**Proof:** By Lemma 14, there exists a dag $\mathcal{G}' \approx \mathcal{G}$ for which $C_{X \cup \{y\}}(\mathcal{G})$ is a covered clique. Let $\mathcal{G}''$ be identical to $\mathcal{G}'$, except that the edges in $C_{X \cup \{y\}}(\mathcal{G})$ are oriented to be in the opposite direction of the corresponding edges in $\mathcal{G}$. By Lemma 13 it follows that $\mathcal{G}'' \approx \mathcal{G}$. □

We can now prove Lemma 3 and Lemma 4. We restate both lemmas below, using the notation developed in this section.

**Lemma 3** *Let $\mathcal{G}$ be any dag and let $C_{\{x,y,z\}}(\mathcal{G})$ be any clique of size three. If any two of the edges in $C_{\{x,y,z\}}(\mathcal{G})$ are in $R_{\mathcal{G}}$, then the third one is also.*

**Proof:** Assume that exactly two of the edges are in $R_{\mathcal{G}}$. Without loss of generality, assume that the edge $z \rightarrow x$ is not in $R_{\mathcal{G}}$ . Let $\mathcal{G}'$ be any dag that includes the edge $x \rightarrow y$. Because $\mathcal{G}'$ is acyclic, $z \rightarrow y$ is in $\mathcal{G}'$. Because $x \rightarrow y$ and $z \rightarrow y$ are both both in $R_{\mathcal{G}'}$, it follows from Corollary 2 that every edge in $C_{\{x,y,z\}}(\mathcal{G}')$ is in $R_{\mathcal{G}'}$, contradicting the assumption that $z \rightarrow x$ is not in $R_{\mathcal{G}}$. □

**Lemma 4** *Let $\mathcal{G}$ be any dag, and let $x \rightarrow y$ be any edge in $\mathcal{G}$ such that $\Pi_y \subseteq \Pi_x \cup x$. The edge $x \rightarrow y$ is in $R_{\mathcal{G}}$ if and only if for every edge $z \rightarrow x$ such that $z$ and $y$ are not adjacent, $z \rightarrow x$ is in $R_{\mathcal{G}}$.*

**Proof: (if)** Assume that for every edge $z \rightarrow x$ such that $z$ and $y$ are not adjacent, $z \rightarrow x$ is in $R_{\mathcal{G}}$. We now show it follows that $x \rightarrow y$ is reversible.

Let $\{z_1 \rightarrow x, \ldots, z_k \rightarrow x\} \subseteq R_{\mathcal{G}}$ be the set of all reversible edges incident into $x$ in $\mathcal{G}$, and let $Z = \{z_1, \ldots, z_k\}$ be the set of tails of these edges. Let $Z_y \subset Z$ be the subset of nodes in $Z$ that are parents of $y$ (see Figure 4a).

By Lemma 14, there exists a dag $\mathcal{G}' \approx \mathcal{G}$ for which $C_{Z \cup \{x\}}(\mathcal{G}')$ is a covered clique (see Figure 4b). By assumption, for every edge $z \rightarrow x$ such that $z$ and $y$ are not adjacent, $z \in Z \setminus Z_y$. As a result, Lemma 13 guarantees that by choosing an appropriate total ordering on the nodes in $Z \cup \{x\}$, we can construct a dag $\mathcal{G}'' \approx \mathcal{G}'$ such that the only edges incident into $x$ are those that have tails in $Z_y$ (see Figure 4c).

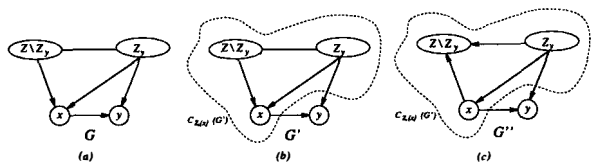

Figure 4: Relevant dags for the proof of Lemma 4

Consider the sequence $\mathcal{S}(\mathcal{G}, \mathcal{G}'')$. By Lemma 10, every edge in $\delta_x(\mathcal{G}, \mathcal{G}'')$ will come before any edge from $\delta_y(\mathcal{G}, \mathcal{G}'')$. Let $\mathcal{G}_i \in \mathcal{D}(\mathcal{G}, \mathcal{G}'')$ be the dag that results after reversing the last edge in $\delta_x(\mathcal{G}, \mathcal{G}'')$. It follows that $\Pi_x^{\mathcal{G}_i} = Z_y$. Furthermore, no edge incident into $y$ has been reversed which implies $\Pi_x^{\mathcal{G}_i} = Z_y \cup x$. Consequently, $x \rightarrow y$ is a covered edge in $\mathcal{G}_i$.

**(Only if)** Let $z \rightarrow x$ be any compelled edge such that $z$ is not adjacent to $y$. It follows immediately that $x \rightarrow y$ is compelled because any dag with $x \rightarrow y$ reversed will contain the v-structure $(z, x, y)$ that is not in $\mathcal{G}$. □